\documentclass[11pt]{article}
\usepackage[preprint]{acl}
\usepackage{times}
\usepackage{latexsym}
\usepackage[T1]{fontenc}
\usepackage[utf8]{inputenc}
\usepackage{microtype}
\usepackage{inconsolata}
\usepackage{amsmath,amssymb}
\usepackage{booktabs,tabularx,multirow}
\usepackage{xcolor}
\usepackage{graphicx}
\usepackage{tikz}
\usetikzlibrary{positioning,fit,arrows.meta}

\newcommand{\folio}{\textsc{Folio}}
\newcommand{\scan}{\texttt{scan}}
\newcommand{\ctx}{\texttt{ctx}}
\newcommand{\dossier}{\texttt{dossier}}

\hypersetup{pdftitle={Folio: Recoverable Multi-Resolution Context Folding for Long-Horizon Language Agents}}
\title{Folio: Recoverable Multi-Resolution Context Folding\\for Long-Horizon Language Agents}
\author{Yubo Ruan$^{1}$ \quad Tao Peng$^{2}$ \quad Jiao Zhang$^{1}$ \quad Chuanguo Ma$^{2}$ \quad Lei Xu$^{2}$ \\ \bfseries
Han Zhou$^{2}$ \quad Wenwu Wang$^{2}$ \quad Xiafei Lv$^{2}$ \quad Xingjie Zhou$^{2}$ \quad Guiju Fan$^{2}$ \\ \bfseries
Hao Wang$^{1}$ \quad Fenghui Zhang$^{2}$ \quad Mao Miao$^{2}$ \quad Shuo Li$^{2}$ \\[4pt]
\normalsize $^{1}$Beijing University of Posts and Telecommunications \qquad $^{2}$Baidu, Inc. \\[3pt]
\small\ttfamily \{ruanyubo, jiaozhang, h.wang\}@bupt.edu.cn \\
\small\ttfamily \{pengtao04, machuanguo, xulei25, zhouhan06, wangwenwu01, lvxiafei, \\
\small\ttfamily zhouxingjie01, fanguiju, zhangfenghui, miaomao01, lishuo02\}@baidu.com}

\begin{document}
\maketitle
\begin{abstract}
Long-horizon language agents accumulate tool results and replies until the history no longer fits the context window. Current agents shrink this history by having a model summarize it, so each compaction rereads and rewrites the history it replaces, and cache-aware triggers decide when to compact while taking the size of the result as given. We present \folio{}, a context manager that records, folds, and recovers history through one set of handles. While the agent runs, \folio{} records three linked levels: a short turn note the model writes as part of its ordinary reply (L1), an event manifest of every call written by the runtime (L2), and the captured source records (L3). Because these views already exist, the runtime folds the context programmatically, selecting coarser views for older turns. The fold prints the handles that index L2 and L3, so recovery tools that read bounded pages, compute over full text in files, and keep findings in a persistent dossier reach the evidence the agent needs with little added input. The fold trigger prices an explicit landing point that includes the retained budget. On 75 LOCA-bench tasks with DeepSeek-V4-Pro behind every system, \folio{} solves 66 tasks against 54 for the same runtime with model-written compaction and 49--59 for Claude Code, Codex, and SoL-Pi, in about half their working time, at 2.4 times lower cost than that runtime; each fold takes a median of 0.6 seconds.
\end{abstract}

\section{Introduction}
\label{sec:intro}
A language agent that solves a long task alternates between generating an action, running a tool, and reading the result. Every result, reply, and intermediate decision is appended to the history that the next request carries. On hard tasks the history outgrows the context window well before the task ends, and even before that point it is reread on every request at the provider's input price. The agent must therefore repeatedly decide what stays in the window, while keeping access to evidence that may matter many steps later.

Existing work addresses parts of this problem (\S\ref{sec:related}). Compaction methods shorten the history by masking or truncating it by rule, or by having a model summarize it~\citep{masking,cliff,acon,contextfolding,agentfold}. Hierarchical and recoverable memory keeps the source reachable below the reduced view: HiAgent summarizes each completed subgoal and retrieves its trajectory on request~\citep{hiagent}, LCM builds a DAG of model-written summaries whose leaves point to the original messages~\citep{lcm}, and SoL-Pi archives large tool outputs behind stable handles~\citep{solpi}. Cache-economic models ask when a reduction pays for itself under prompt caching: Context Is Inventory treats a session as a rent-or-buy problem in which carrying context costs cached reads and a restart costs a rebuild~\citep{inventory}, and SoL-Pi's Online Context Compact compares projected savings with the cache-rewrite cost before invoking the base agent's summarizer~\citep{solpi}.

Two gaps remain across these lines. First, the reduced representation is produced at compaction time, usually by a model. Each compaction reads the material it replaces and writes a new object, so the work grows with the number of compactions, and the link between the new object and its source is whatever the compaction step recorded. Second, the timing models take the post-reduction size as given. In the restart model of Context Is Inventory the context returns to its working floor after every rebuild, so the landing point is not a decision, and a cost gate that decides whether to compact leaves the size of the result to the summarizer. For folding, which keeps part of the history, how much is kept is a choice with its own cost, and the trigger should price it.

We present \folio{}, a context manager in which the history representation is formed during execution and folding becomes a selection among views that already exist. \folio{} \emph{records} the history as three linked levels while the agent runs: a turn note of a few lines that the model writes as part of its ordinary reply (L1), instead of a summary produced by a separate request; an event manifest the runtime writes for every call, reply, and turn (L2); and the source records of the captured event log (L3). Every call has one handle, such as \texttt{t3.12}, that stays valid across levels and across folds. Because the views already exist, \folio{}'s runtime \emph{folds} the context programmatically: it selects coarser views for older turns and keeps the open turn's earliest small results and newest results verbatim. The fold prints the same handles that index L2 and L3, so the agent can \emph{recover} the evidence it actually needs quickly and with little added input: \scan{} reads a bounded part of a source, \ctx{} computes over its full text in files, and \dossier{} keeps findings that persist across folds. The fold trigger prices the landing point $S=B+L+D$, the stable prefix plus the retained layout budget plus the rendered dossier, so the retained size enters the cost decision. These parts are not independent features: one handle space and one renderer tie the representation, the fold, recovery, and cost control together.

This paper makes four contributions:
\begin{itemize}
\setlength\itemsep{0pt}
\item An execution-time, multi-resolution history in which the identity of a historical object is separate from the resolution at which it is shown, with byte-offset and digest addressing into the event log (\S\ref{sec:history}).
\item A programmatic fold in which the runtime selects and demotes existing views and never re-renders an already sent line at a richer level, together with three recovery operations that share the fold's handles and so reach folded evidence precisely and cheaply (\S\ref{sec:folding}, \S\ref{sec:access}).
\item A landing-priced fold trigger that accrues carried excess against each request's own landing point, so it needs neither a fixed fold interval nor linear context growth (\S\ref{sec:trigger}).
\item An evaluation on 75 LOCA-bench tasks run as five 15-task sessions, against Claude Code, Codex, SoL-Pi, and the same runtime without \folio{}, all on DeepSeek-V4-Pro, with component ablations, a landing-budget sweep, a check of every fold, and a recovery case (\S\ref{sec:results}).
\end{itemize}

With the same model behind every agent, \folio{} solves 66 of 75 tasks, against 54 for the same runtime with model-written compaction, 59 for Claude Code, 57 for Codex, and 49 for SoL-Pi, and needs a mean of 10.2 minutes per task against 16.3--21.0. Against the same runtime it costs 2.4 times less in normalized tokens. Its folds are completed by the runtime alone in a median of 0.6 seconds, against 101 seconds for a model-written compaction. Removing the outline from the fold nearly doubles the number of folds and raises cost by 63\%, and a sweep of the landing budget shows that the retained size trades folds and recoveries against carried context.

\begin{figure*}[t]
\centering
\includegraphics[width=\textwidth]{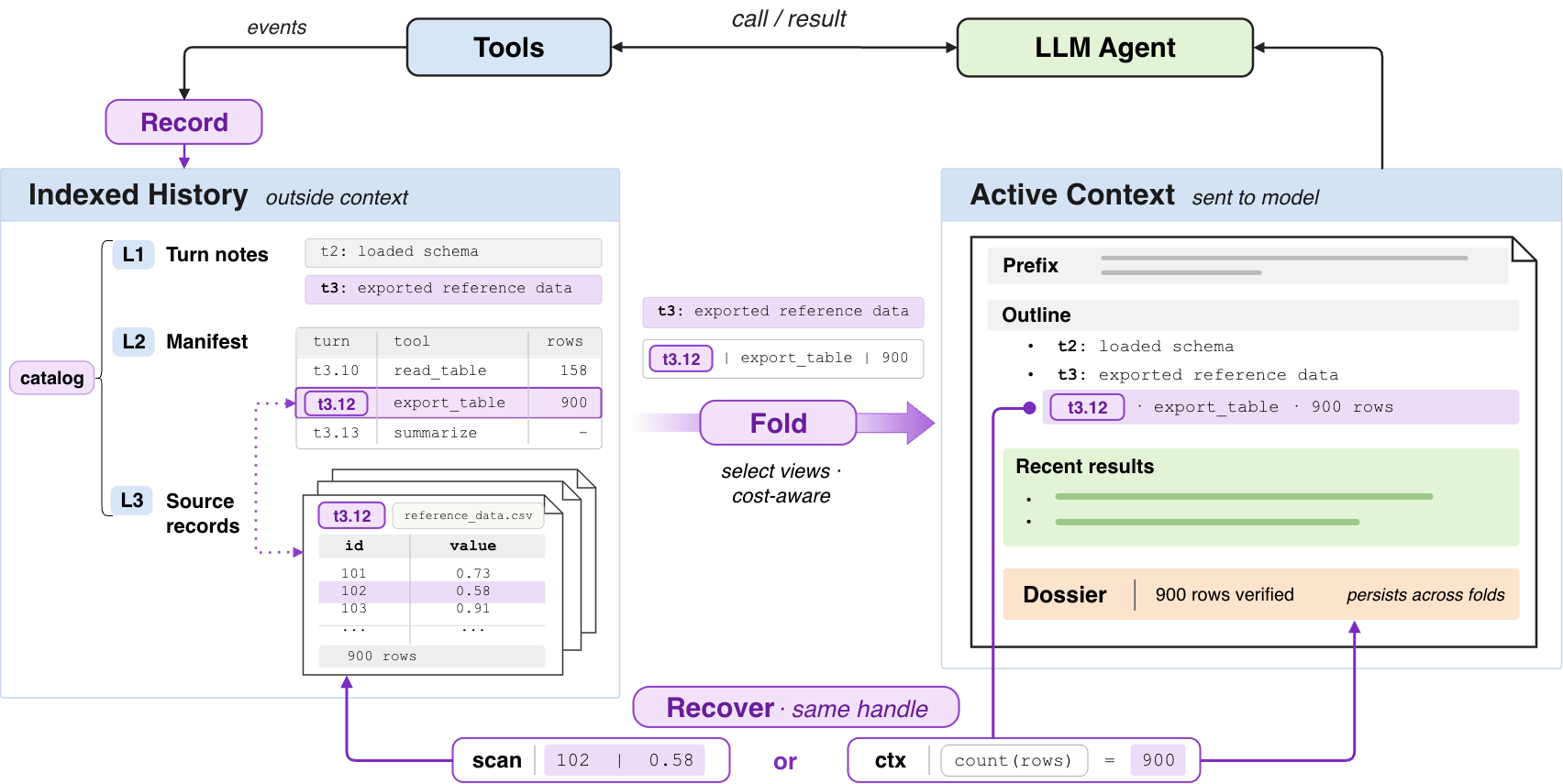}
\caption{Overview of \folio{}. \textbf{Record}: the runtime turns tool events into a three-level indexed history. \textbf{Fold}: the cost-aware trigger selects views of this history to build the active context. \textbf{Recover}: the handle in the outline (\texttt{t3.12}) is passed to \scan{} for a bounded read or to \ctx{} to compute over the full text, and the result is kept in the dossier.}
\label{fig:overview}
\end{figure*}

\section{Problem Formulation}
\label{sec:problem}

\paragraph{Sessions and contexts.}
A session consists of turns. A turn opens with a user message (or a runtime continuation) and may contain many model requests and tool calls before it closes. We index model requests by $t$, because each request is a unit of both context construction and billing. Let $\mathcal{H}_t$ be the history captured before request $t$ and $X_t$ the context actually sent. A context manager constructs $X_t$ from $\mathcal{H}_t$ subject to $|X_t|+r_t\le W$, where $W$ is the window and $r_t$ the space reserved for the response. The history is a record of past observations, not the current environment: rerunning a query may return different data, and rerunning a write may repeat its side effects. Recovering history therefore means reading what was captured, not calling the tool again.

\paragraph{Cost.}
Prompt-cached APIs bill cached input, uncached input, cache creation, and output at different prices. For request $t$ let $h_t$, $u_t$, $w_t$, and $o_t$ be these mutually exclusive token counts and $p^h,p^u,p^w,p^o$ their prices. The cost of a run is
\begin{equation}
\label{eq:cost}
C=\sum_{t\in\mathcal{R}}\bigl(p^h h_t+p^u u_t+p^w w_t+p^o o_t\bigr),
\end{equation}
where $\mathcal{R}$ contains every billed request: the agent's own, any auxiliary model calls for notes or memory maintenance, and retries. Tokens read back from folded history are already inside these requests and are not counted again. Dividing by $p^h$ expresses all terms in cached-read equivalents while keeping the price ratios $\rho_u=p^u/p^h$, $\rho_w=p^w/p^h$, and $\rho_o=p^o/p^h$.

\paragraph{Three decisions.}
A context manager makes three decisions. It chooses a \emph{representation} of history that can be shown at less than full size. It chooses \emph{when} to replace the resident history with a smaller view. It chooses \emph{how large} that view is. Under Eq.~\ref{eq:cost} the last two interact: a fold breaks the cached prefix and the next request pays $\rho_u$ or $\rho_w$ on what is rebuilt, while every later request pays $p^h$ on what is kept. Keeping less also changes what follows, because the agent may search, reread, or reason again to recover what it no longer sees.

\paragraph{Restart and fold.}
The rent-or-buy model of \citet{inventory} analyzes \emph{restarts}. After a restart the context returns to the working floor $L_{\mathrm{floor}}$ (system prompt, tool definitions, project instructions), and the rebuild cost is priced on that floor. The post-restart size is therefore fixed by construction. A \emph{fold} keeps part of the history, so the context after it lands at
\begin{equation}
\label{eq:landing-point}
S=B+L+D,
\end{equation}
where $B$ is the stable prefix, $L$ the layout budget for the folded history and the open turn, and $D$ the dossier of retained findings (\S\ref{sec:access}). $B$ plays the role of $L_{\mathrm{floor}}$, but $L$ is a parameter of the policy and $D$ changes as the agent works. A trigger for folding has to price this landing point rather than assume it.

\paragraph{Recoverability contract.}
We call a fold \emph{recoverable} if, for every object that was captured, is still stored, and has a valid handle, three properties hold after any number of folds: (i) \emph{retention}, the fold does not delete the source record; (ii) \emph{resolution}, the handle still resolves to the same record; and (iii) \emph{exact read}, a read through the handle returns that record's text, or reports that only a cut copy is available. The contract says nothing about whether the agent asks for the right record or uses it correctly; those are measured as task outcomes (\S\ref{sec:results}). It also does not cover content the tool never returned or the runtime never captured.

\section{Folio}
\label{sec:folio}

\begin{figure*}[t]
\centering
\includegraphics[width=\textwidth]{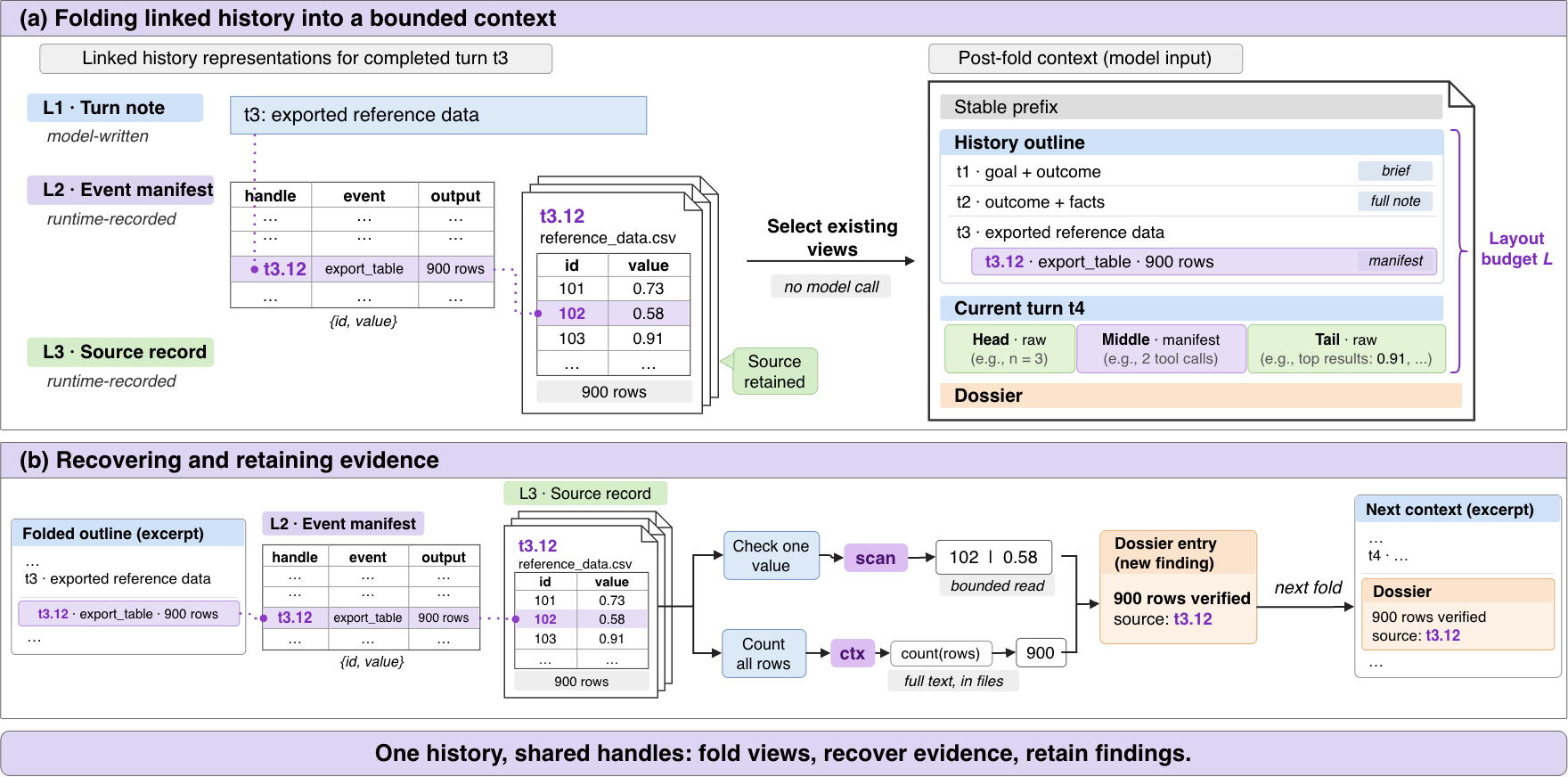}
\caption{Folding and recovery on one example. (a) Turn \texttt{t3} exists as a turn note (L1), the manifest row of call \texttt{t3.12} (L2), and the 900-row table it returned (L3). Within the layout budget $L$, the fold shows \texttt{t1} briefly, \texttt{t2} with its full note, and \texttt{t3} with the row of \texttt{t3.12}; the open turn \texttt{t4} keeps a raw head and tail. (b) Through the same handle, \scan{} checks one value and \ctx{} counts every row; the finding and its source go to the dossier, which the next fold carries.}
\label{fig:fold-recover}
\end{figure*}

\subsection{Overview}
\label{sec:overview}

\folio{} sits between the agent and its tools and runs three operations that repeat throughout a session (Figure~\ref{fig:overview}). \emph{Record}: as calls and results pass between the agent and its tools, the runtime appends each event to the source records (L3), writes an event-manifest row for each call and reply as it arrives (L2), and stores the model's turn note when the turn closes (L1). Together these form an indexed history outside the context (\S\ref{sec:history}). \emph{Fold}: when the trigger fires (\S\ref{sec:trigger}), the runtime selects views of these rows within the layout budget and builds the active context: the stable prefix, an outline of earlier turns, the open turn's recent results, and the dossier (\S\ref{sec:folding}). \emph{Recover}: the outline prints call handles; the agent passes a handle to \scan{} or \ctx{} to reach the folded source, and keeps what it learns in the dossier (\S\ref{sec:access}). New calls extend the same levels. Figure~\ref{fig:fold-recover} follows one call, \texttt{t3.12}, through all three operations.

\subsection{Multi-Resolution History}
\label{sec:history}

The three levels have different authors and different jobs (Figure~\ref{fig:fold-recover}a, left). The model writes only L1; the runtime writes L2 and L3. None is a complete encoding of the level below: a coarse view can omit a value, and that value stays reachable through the handle.

\paragraph{L1: turn notes.}
The model ends a turn's reply with a short note block containing an \texttt{outcome} line and any number of \texttt{fact} lines (user constraints, choices and their reasons). The runtime parses the block out of the user-visible reply and stores it with the turn's status, goal, and raw size. A turn without a note falls back to the head of its final reply. Writing notes is part of ordinary generation, not a separate request.

\paragraph{L2: event manifest.}
A \emph{tool row} records the tool, an argument digest, the purpose the model gave for the call, status, output size, the first content line, and for results above 1K tokens a one-line shape and sample of the parsed structure (for example \texttt{array[135] of \{id:number,name:string,\ldots\}}). A \emph{say row} records text the model wrote between calls and where it fell in the call sequence. A \emph{turn row}, written when the runtime declares the turn closed, records the goal, the reply head, tool counts, and the rendered size of the turn at every level. Tool and say rows of an open turn are written as their events land, so the manifest is current at any fold.

\paragraph{L3: source records.}
The append-only event log holds user messages, tool requests, routed results, and assistant messages. When the runtime cut a result before routing it to the model, the complete output is kept in a separate file referenced from the event.

\paragraph{Identity and addressing.}
A call is named \texttt{t$N$.$k$}: the $k$-th tool row of turn $N$ in the order results settled, which is also the order \scan{} resolves. Ranges are \texttt{t$N$.$a$-$b$}. Each tool row stores the byte range of its routed event in the log and a digest of those bytes. A read seeks to the range and checks the digest; a mismatch, a missing file, or a missing range falls back to parsing the log. The offset is thus an access hint and the handle is the identity. L1 rows point to the byte range of their turn's L2 rows. A result that settles after its turn closed is appended later and listed in an index, so the handle still finds it. When the manifest file grows past a size limit or a day boundary, it is split at a turn boundary into archived segments with logical offsets, and an interrupted split is completed on the next load.

\subsection{Folding as View Selection}
\label{sec:folding}

\paragraph{Layout.}
A fold renders into a layout budget $L$ (the landing budget of Eq.~\ref{eq:landing-point}; Figure~\ref{fig:fold-recover}a, right). An outline share $\alpha$ holds the reading guide, the closed turns, and the rows of the open turn; the rest holds raw results of the open turn, split into a head share $\beta$ and a tail:
\begin{equation}
\label{eq:layout}
\begin{aligned}
L_{\mathrm{out}}&=\alpha L,\qquad L_{\mathrm{head}}=\beta(1-\alpha)L,\\
L_{\mathrm{tail}}&=(1-\beta)(1-\alpha)L.
\end{aligned}
\end{equation}
We use $\alpha=\beta=0.2$. The budgets are planning targets: rows are discrete, and the raw area is further capped so that the post-fold context leaves headroom below the window.

\paragraph{Closed turns.}
The planner draws each closed turn at one of three levels: \emph{l2} (head line, outcome and facts, one row per call), \emph{l1\_full} (the same without call rows), or \emph{l1\_brief} (goal cut to 80 characters and the outcome). The planner starts with the newest turn at l2, the two before it at l1\_full, and older turns at l1\_brief; in Figure~\ref{fig:fold-recover}a, \texttt{t1} is brief, \texttt{t2} keeps its full note, and \texttt{t3} keeps the manifest row of \texttt{t3.12}. If these floors exceed the budget, it demotes from the oldest turn; it then upgrades from the newest turn while the increase fits. A turn whose full call list does not fit is tried at l2 with its oldest calls folded into one paging row (\texttt{t$N$.1-$m$ ×$m$ ok~$x$ failed~$y$: tools}) before it settles for l1\_full. Beyond the newest 30 turns, older turns merge into phase rows of 10 turns that keep each turn's note lines.

\paragraph{Monotone demotion.}
The planner reads the levels recorded by the previous fold and caps each turn at its previous level and paging. A line that was sent once is never redrawn richer, even when a later fold has room. Repeated folds therefore only shorten the carried outline, and the part of it that does not change stays a byte-identical prefix for the provider's cache.

\paragraph{The open turn.}
The open turn is laid out from its rows (\texttt{t4} in Figure~\ref{fig:fold-recover}a). The head takes the turn's earliest results of at most 1K tokens in call order until one does not fit; these are typically the task's inputs. The tail takes results from the newest call backward until one does not fit. Every other call keeps its row, and a large one also keeps its structure line. If the rows still exceed the outline area, the oldest middle rows fold into a paging row. When the open turn continues across several folds, the next fold parses the block the previous fold sent, appends the new calls, and lays out the whole turn by the same rules with the same call numbers.

\paragraph{A programmatic fold.}
The runtime folds by selecting, cutting, and arranging rows that already exist. After rendering, it rebuilds the closed turns from the L1/L2 files alone, re-renders them, and compares the digest with the text it sent; a mismatch is recorded with the fold. Model work in \folio{} is limited to the L1 notes and dossier entries written during execution, plus one fallback: when the dossier's entries exceed half the window, a low-effort auxiliary request merges duplicate entries in chunks, and a chunk that fails stays as it was. All of these are billed requests in Eq.~\ref{eq:cost}.

\subsection{Access Through Shared Handles}
\label{sec:access}

The outline tells the agent where evidence is; three operations bring it back (Figure~\ref{fig:fold-recover}b). All three take the handles the outline prints, so a read lands on exactly the object the fold removed.

\paragraph{\scan{}: locate and read.}
\scan{} is read-only. \texttt{list} expands a handle: a turn into its call rows, a range into rows one page at a time, a long call result into a structure map of line-numbered headings, signatures, or paragraph heads. The map is returned only when it is informative (at least three entries and one per 40 lines); otherwise the first text page comes back. \texttt{read} returns a bounded page of a call's text by line range, by \texttt{grep}, or centered on a hit, at most 200 lines and 8K tokens per answer, and names the next page. \texttt{find} searches all folded results and replies and returns one line per hit with its handle, or snippets, rows, or whole calls, with a cursor to continue. Patterns are plain text with optional anchors, so a search cannot stall the runtime. When the routed copy of a result was cut, \texttt{read} pages the complete output file instead and says so if that file is missing or too large. \scan{} does not read or search the results of earlier \scan{} calls: a read of such a call returns an error that names the original handle, and \texttt{find} skips them, so a search does not count the same hit twice or match its own earlier pages.

\paragraph{\ctx{}: compute in files.}
Some questions need the whole of a result: counting rows, joining two lists, or checking every entry against a condition. Paging such a result into the window costs input on every page and invites transcription errors. \ctx{} binds the full text of a handle or range to a file, runs a shell or Python command the agent writes over it, and returns a bounded answer. Its output can be saved as \texttt{derived/<name>}, which later \ctx{} calls take as input and which the dossier lists.

\paragraph{\dossier{}: keep what the task needs.}
The dossier is the agent's store of working results. A \emph{finding} holds one fact under a key, with the handle it came from; a \emph{note} holds a judgment and is append-only; a \emph{rule} holds a constraint that must survive turns. Entries have stable ids (\texttt{m12}) and are rewritten by key or forgotten by id; notes cannot be forgotten. Each entry is written either for the current turn (the default) or as long-term. Current-turn entries are shown only while that turn runs; long-term entries, mostly rules, persist across turns. Separately, the runtime logs the user's stated requirements verbatim and shows them in every turn. Every fold appends the dossier after the outline under its own budget (12\% of the window). Within that budget, the visible notes are always shown and the newest 30 file pointers are listed; the remaining budget goes to long-term entries and then to the current turn's findings, newest first in each group, so older entries are the first to be omitted. Entries new since the last fold are marked, and the agent's latest plan in this turn is attached at the top together with the handles it has already read back in this turn. The agent can therefore resume from its unfinished steps without rereading what it already extracted.

\paragraph{Example.}
Figure~\ref{fig:fold-recover}b continues the example. Call \texttt{t3.12} exported a 900-row table; after the fold, the outline keeps only its manifest row \texttt{t3.12 $\cdot$ export\_table $\cdot$ 900 rows}. To check one value, the agent reads \texttt{t3.12} with \scan{} and a \texttt{grep} for the id, and receives a bounded page with the row \texttt{102 | 0.58} instead of the table. To confirm that all 900 rows are present, it runs \ctx{} over the same handle with a row count; the count runs over the full text in a file and returns only \texttt{900}. The agent then writes the finding with its source \texttt{t3.12} to the dossier. The next fold carries neither the table nor the reads, and the finding stays in the context with the handle to check it again.

\subsection{Landing-Priced Fold Trigger}
\label{sec:trigger}

The trigger decides when to fold; the layout budget $L$ decides where the fold lands. Following the rent-or-buy view~\citep{inventory}, a fold is worth it once the cached reads carried above the landing point have cost as much as rebuilding the landing point.

\paragraph{Landing point.}
The stable prefix $B$ is measured once: the provider-counted input of the session's first request minus the estimated tokens of its message text. After every response, the runtime renders the dossier exactly as the next fold would and measures its size $D_t$. The landing point and the rebuild cost of a fold decided at request $t$ are
\begin{equation}
\label{eq:landing}
S_t=B+L+D_t,\qquad K_t=\rho_w\,(L+D_t),
\end{equation}
since a fold rewrites everything above the stable prefix.

\paragraph{Carried excess.}
For each settled request with provider-counted cached input $h_t$, the excess is charged against that request's own landing point:
\begin{equation}
\label{eq:excess}
E_t=E_{t-1}+\max(0,\,h_t-S_t).
\end{equation}
A request that missed the cache has $h_t$ below $S_t$ and adds nothing. The runtime folds when $E_t\ge\lambda K_t$, judged after a response and before its tool calls run; a committed fold resets $E$. A separate safety line folds whenever the context reaches a fixed share of the window, and the layout keeps a minimum headroom above the landing point.

\paragraph{What the rule does not assume.}
Eq.~\ref{eq:excess} sums observed requests. It does not need the number of requests between folds, a constant growth rate, or a fixed post-fold size: when the dossier grows, $S_t$ and $K_t$ move with it, and the interval between folds changes accordingly. The landing point includes the retained budget $L$, which a restart model sets to zero by construction. Two costs are left out of $K_t$ on purpose: the extra reasoning after a fold and the tokens the agent reads back. Neither is known when the decision is made; both appear in later requests, where Eq.~\ref{eq:excess} and Eq.~\ref{eq:cost} count them. Our reference configuration uses $L=50$K, $\lambda=1$, and DeepSeek-V4-Pro's price ratios ($\rho_u=\rho_w=30$, $\rho_o=90$).

\section{Experimental Setup}
\label{sec:setup}

\paragraph{Tasks.}
We use LOCA-bench~\citep{locabench}, an open benchmark of tool-using tasks whose context grows with the environment. It has 15 environments (course scheduling from Canvas announcements and email, inventory and sales analysis over WooCommerce, invoice checking, A/B-test analysis, and others), each with five seeds, for 75 tasks with deterministic graders. We build five \emph{chains}: chain $k$ takes the $k$-th seed of every environment, and its 15 tasks are given to the agent one after another in a single session, so later tasks run on top of the history of earlier ones. Each system runs each chain once, giving 75 graded runs per system on identical task sequences (the task prompts match by hash across systems). Under \folio{}, 60 of the 75 tasks trigger at least one fold, and the median context at a fold is 242K tokens.

\paragraph{Systems.}
All systems use DeepSeek-V4-Pro. \folio{} runs in the Piki runtime with a 1M-token window and the reference configuration of \S\ref{sec:trigger} ($L=50$K, $\lambda=1$). We also run \folio{} with a 128K window, where the window cap cuts the landing budget to about 16.5K. Piki-Classic is the same runtime without \folio{}, which summarizes the history with the model when the context reaches 90\% of the window. Claude Code, Codex, and SoL-Pi (all four mechanisms enabled;~\citealp{solpi}) run with a 1M-token window. No system has a step or time budget beyond the benchmark's own timeout. Ablations (\S\ref{sec:exp-ablation}) use the same 1M configuration; the landing-budget sweep (\S\ref{sec:exp-landing}) runs the fifth chain twice per budget, once in ascending and once in descending order of $L$.

\paragraph{Metrics.}
\emph{Success} is the number of the 75 runs the grader accepts; we also report the mean and standard deviation of successes per chain (out of 15). \emph{Time} is the agent's working time per task in minutes, excluding environment setup and grading. For \folio{} and Piki-Classic, whose runtime records provider usage, we report tokens per task split into cached input, uncached input, and output, and a normalized cost in cached-read equivalents, $h+30u+90o$, using DeepSeek-V4-Pro's price ratios (Eq.~\ref{eq:cost}). The closed systems do not expose provider usage, so we compare them on success and time only. Paired comparisons over the 75 tasks use an exact McNemar test for success. The benchmark's contamination scan flags runs that read grader data; flagged runs are kept and their counts are listed in Table~\ref{tab:main}.

\section{Results and Analysis}
\label{sec:results}

We report five experiments: a comparison with other agents (\S\ref{sec:exp-main}), component ablations (\S\ref{sec:exp-ablation}), a sweep of the landing budget (\S\ref{sec:exp-landing}), the cost and correctness of the fold itself (\S\ref{sec:exp-fold}), and a recovery case (\S\ref{sec:exp-case}).

\subsection{Comparison with Other Agents}
\label{sec:exp-main}

\begin{table}[t]
\centering\small
\setlength{\tabcolsep}{3pt}
\begin{tabular}{@{}lccccc@{}}
\toprule
& \multicolumn{2}{c}{Success} & \multicolumn{2}{c}{Time (min/task)} & \\
\cmidrule(lr){2-3}\cmidrule(lr){4-5}
System & /75 & per chain & mean & median & Flag. \\
\midrule
Claude Code & 59 & 11.8\,$\pm$\,0.8 & 18.5 & 13.7 & 9 \\
Codex & 57 & 11.4\,$\pm$\,1.7 & 20.2 & 15.4 & 27 \\
SoL-Pi & 49 & \phantom{0}9.8\,$\pm$\,1.3 & 21.0 & 17.9 & 24 \\
Piki-Classic & 54 & 10.8\,$\pm$\,1.3 & 16.3 & 13.7 & 11 \\
\midrule
\folio{} (128K) & 60 & 12.0\,$\pm$\,1.2 & 16.4 & \phantom{0}7.8 & 9 \\
\folio{} (1M) & \textbf{66} & \textbf{13.2\,$\pm$\,1.5} & \textbf{10.2} & \textbf{\phantom{0}6.9} & 8 \\
\bottomrule
\end{tabular}
\caption{LOCA-bench, 5 chains $\times$ 15 tasks, DeepSeek-V4-Pro for every system. Per chain: mean $\pm$ s.d.\ of successes out of 15. Flag.: runs flagged by the benchmark's contamination scan.}
\label{tab:main}
\end{table}

\begin{figure*}[t]
\centering
\includegraphics[width=\textwidth]{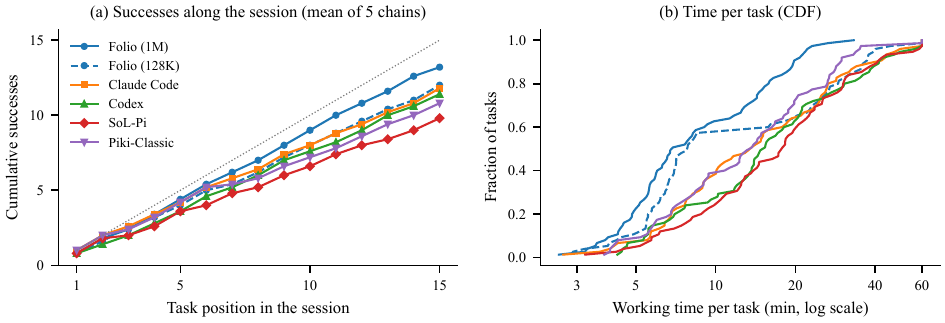}
\caption{(a) Cumulative successes along the 15-task session, averaged over the five chains; the dotted line solves every task. (b) Distribution of working time per task over all 75 tasks.}
\label{fig:main}
\end{figure*}

\paragraph{Against the same runtime.}
Piki-Classic differs from \folio{} only in context management. It solves 54 of 75 tasks against \folio{}'s 66; paired by task, \folio{} solves 14 tasks that Piki-Classic fails and fails 2 that it solves (exact McNemar $p=0.004$). \folio{} is faster on 57 of 75 tasks, with a median per-task time ratio of 1.45.

\begin{figure}[t]
\centering
\includegraphics[width=0.92\columnwidth]{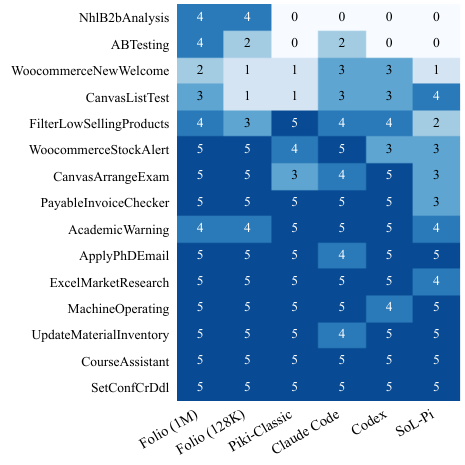}
\caption{Successes per LOCA-bench environment (out of 5 seeds). Rows are sorted by the mean of Claude Code, Codex, and SoL-Pi, hardest first.}
\label{fig:env}
\end{figure}

\paragraph{Against other agents.}
\folio{} also solves more tasks than Claude Code (59), Codex (57), and SoL-Pi (49) (Table~\ref{tab:main}, Figure~\ref{fig:main}a). Paired by task, it solves 14 tasks that Claude Code fails and fails 7 that Claude Code solves ($p=0.19$); against Codex the split is 14 to 5 ($p=0.06$) and against SoL-Pi 20 to 3 ($p<0.001$). With 75 tasks, the differences from Piki-Classic and SoL-Pi are significant at 0.05 and those from Claude Code and Codex are not. Figure~\ref{fig:main}a shows that \folio{}'s lead widens in the second half of the session, when the history of earlier tasks has been folded several times. Figure~\ref{fig:env} breaks the result down by environment. The gap is concentrated in two environments in which every other system fails most seeds: NHL B2B analysis (\folio{} 4/5, every other system 0/5) and A/B-test analysis (4/5 against at most 2/5).

\folio{} is also faster (Figure~\ref{fig:main}b). Its mean working time is 10.2 minutes per task, against 16.3 for Piki-Classic, 18.5 for Claude Code, 20.2 for Codex, and 21.0 for SoL-Pi. It takes less time than Claude Code on 61 of 75 tasks, less than Codex on 65, and less than SoL-Pi on 64; the median per-task ratio of the baseline's time to \folio{}'s is 1.56, 1.80, and 1.90. With the window cut to 128K, \folio{} still solves 60 tasks, more than any baseline run with a 1M window, at 16.4 minutes per task.

\begin{table}[t]
\centering\small
\setlength{\tabcolsep}{3pt}
\begin{tabular}{@{}lccc@{}}
\toprule
& \multicolumn{2}{c}{\folio{}} & Piki- \\
\cmidrule(lr){2-3}
Per task & 1M & 128K & Classic \\
\midrule
Input tokens (M) & 4.01 & 2.72 & 14.12 \\
\quad cached & 3.69 & 2.46 & 13.29 \\
\quad uncached & 0.32 & 0.27 & 0.83 \\
Output tokens (M) & 0.07 & 0.10 & 0.10 \\
Cache hit rate & 92.0\% & 90.2\% & 94.1\% \\
Normalized cost (M) & \textbf{19.75} & 19.83 & 47.00 \\
Folds / compactions & 1.13 & 3.15 & 0.32 \\
Model requests & 29.5 & 39.9 & 34.4 \\
\bottomrule
\end{tabular}
\caption{Usage per task (means over 75 tasks). Normalized cost is $h+30u+90o$ in cached-read token equivalents; for Piki-Classic it includes its summary requests.}
\label{tab:usage}
\end{table}

\paragraph{Cost.}
Table~\ref{tab:usage} compares usage within the Piki runtime, where provider usage is recorded. Piki-Classic's normalized cost is 2.38 times \folio{}'s (3{,}525M against 1{,}481M tokens over 75 tasks). \folio{} is cheaper on 68 of 75 tasks and in all 15 environments, with a median per-task ratio of 2.29. Both agents make about the same number of tool calls (10{,}701 against 10{,}173), so the difference comes from how much history each request carries: Piki-Classic compacts at 90\% of the window, so its requests carry up to about 900K tokens before each compaction, and it reads 3.5 times as much input. Its 24 compactions also cost 420M normalized tokens for the summary requests alone, 28\% of \folio{}'s total cost. Within \folio{}, 92\% of input is served from cache, and the 128K configuration costs the same as the 1M one within 0.4\%: it folds almost three times as often and reads 32\% less input, but it issues 35\% more requests and generates 46\% more output.

\subsection{Ablations}
\label{sec:exp-ablation}

We remove components cumulatively from the full system (A). B removes the dossier; C also removes \scan{} and \ctx{}; D also removes the multi-resolution outline, so the whole layout budget goes to raw text (outline share $\alpha=0$). A separate variant removes only the L1 notes. All variants keep $L=50$K, the 1M window, and the landing-priced trigger.

\begin{table}[t]
\centering\small
\setlength{\tabcolsep}{2.6pt}
\begin{tabular}{@{}lccccc@{}}
\toprule
Variant & $n$ & Succ. & Cost & Folds & Min \\
\midrule
A \folio{} & 75 & 66 & 19.75 & 85 & 10.2 \\
B $-$ \dossier{} & 75 & 65 & 19.73 & 85 & 10.7 \\
$-$ L1 notes & 75 & 63 & 19.43 & 80 & \phantom{0}9.5 \\
\midrule
A \folio{} & 15 & 13 & 18.07 & 16 & \phantom{0}7.8 \\
B $-$ \dossier{} & 15 & 13 & 20.89 & 18 & 12.5 \\
C $-$ \scan{}, \ctx{} & 15 & \phantom{0}8$^\dagger$ & 18.65 & 18 & 21.7 \\
D $-$ outline & 15 & 11 & 29.44 & 31 & 15.5 \\
\bottomrule
\end{tabular}
\caption{Ablations (1M window, $L=50$K). Top: all 75 tasks. Bottom: the first chain, on which C and D were run. Cost: normalized tokens per task (M); Min: minutes per task. Each of B--D removes one more component than the row above. $^\dagger$Two timeouts and one agent error.}
\label{tab:ablation}
\end{table}

\paragraph{Dossier and notes.}
On all 75 tasks, removing the dossier changes one success (66 to 65) and the cost by 0.1\%; removing the L1 notes changes three successes (66 to 63) and lowers the cost by 1.6\%. Running the same configuration on another chain moves its successes by 3 to 4 out of 15 (Table~\ref{tab:main}), so neither difference is distinguishable from that variation on this benchmark.

\paragraph{Recovery tools and outline.}
On the first chain, removing \scan{} and \ctx{} lowers successes from 13 to 8, although three of the seven failures are timeouts or agent errors and this run had 21 provider retries against none for A. Removing the outline as well has the clearest effect on cost. With $\alpha=0$, the layout holds only raw results, so the context refills faster after each fold: D folds 31 times against 16 for A, issues 2.3 times as many requests, and costs 63\% more per task (29.4M against 18.1M). Its success (11 of 15) is within the chain-to-chain variation of A. Keeping the outline thus does not cost budget; it lets each fold retain more of what the agent needs, so fewer folds and fewer requests are required.

\subsection{Landing Budget}
\label{sec:exp-landing}

\begin{figure*}[t]
\centering
\includegraphics[width=\textwidth]{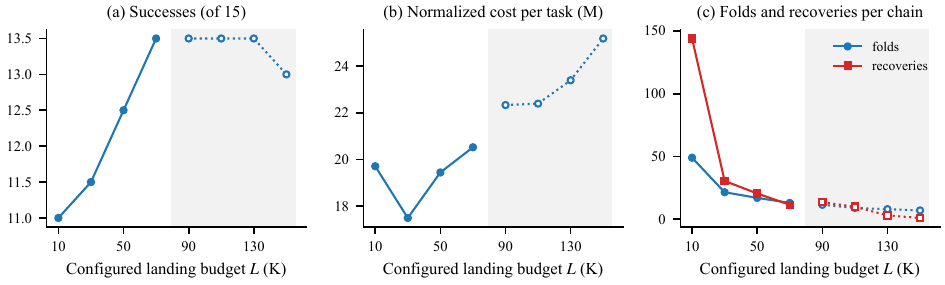}
\caption{Landing-budget sweep on the fifth chain (15 tasks; mean of an ascending and a descending run, one run at $L=10$K). Recoveries are \scan{} and \ctx{} calls that reach content removed by the last fold. In the shaded region the raw area hits its 60K cap, so the effective budget is 78--90K (hollow markers).}
\label{fig:landing}
\end{figure*}

\begin{figure*}[t]
\centering
\includegraphics[width=\textwidth]{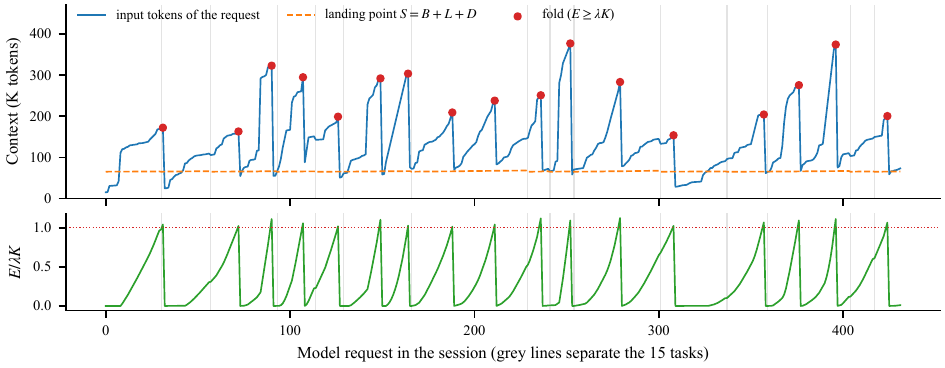}
\caption{One \folio{} session (fifth chain, 1M window, 432 requests). Top: input tokens of each request and the landing point $S$; red dots mark the 17 folds. Bottom: carried excess relative to the fold price; a fold occurs when it reaches 1.}
\label{fig:ledger}
\end{figure*}

\paragraph{The trigger decides the folds.}
With $L=50$K and a 1M window, all 85 folds of the 75 tasks were triggered by the landing-priced rule and none by the safety line. Figure~\ref{fig:ledger} traces one session: the excess $E$ accrues only while the cached context exceeds the landing point, so the interval between folds varies from task to task with the rate at which the context grows, and every fold returns the context close to $S$. On the 128K window the cap on the landing point leaves little room above it, and 201 of 236 folds were triggered by the safety line. The trigger prices a fold at $K=\rho_w(L+D)$; on the 1M runs, $K/\rho_w$ is a median 1.2 times the size the fold actually rendered, so the price of a fold is close to what it rewrites.

\paragraph{Sweep.}
We vary $L$ from 10K to 150K on the fifth chain, running each budget once in ascending and once in descending order (Figure~\ref{fig:landing}). Above 75K the raw area reaches its 60K cap, so configured budgets of 90K--150K take effect as 78K--90K while $K$ is still priced on the configured value; we therefore read the trend from the four budgets that took effect as configured. From 10K to 70K, folds per chain fall from 49 to 13 and recoveries of folded content from 144 to 12, so a smaller landing point is paid for by more folds and more reads back. Successes rise from 11 to 13.5 of 15, though the two runs at one budget differ by up to three tasks (14 and 11 at 50K). Cost is lowest at 30K (17.5M per task) and rises on both sides: at 10K the extra folds and recoveries dominate, and above 30K each request carries more of the landing point. In the capped region cost keeps rising to 25.2M per task at a configured 150K with no further gain in success. The reference $L=50$K lies within 11\% of the lowest cost and close to the highest success.

\subsection{Cost and Correctness of the Fold}
\label{sec:exp-fold}

\begin{table}[t]
\centering\small
\setlength{\tabcolsep}{3pt}
\begin{tabular}{@{}lccc@{}}
\toprule
& \multicolumn{2}{c}{\folio{} fold} & Classic \\
\cmidrule(lr){2-3}
& 1M ($n{=}85$) & 128K ($n{=}236$) & ($n{=}24$) \\
\midrule
Model calls & 0 & 0 & 24 \\
Latency & 613\,ms & 346\,ms & 101\,s \\
Context before & 242K & 121K & 908K \\
Context after & 64K & 39K & 58K \\
Compression & 7.3$\times$ & 7.0$\times$ & -- \\
Dossier & 1.2K & 1.7K & -- \\
Re-render mismatch & 0 & 0 & -- \\
\bottomrule
\end{tabular}
\caption{Folds of the main \folio{} runs and compactions of Piki-Classic. Latency, context sizes, and ratios are medians. Compression is the folded source history over the rendered outline.}
\label{tab:fold}
\end{table}

Table~\ref{tab:fold} summarizes the folds of the main runs. All were completed by the runtime without a model request, and the median fold took 613\,ms on the 1M window. A fold reduced the median context from 242K to 64K tokens, compressing the folded history by a median factor of 7.3. Piki-Classic instead compacts once the context reaches about 908K tokens; each of its 24 compactions is a model request that rereads that history, and the median compaction took 101 seconds, 165 times as long as a fold.

After every fold, the runtime rebuilt the closed turns from the L1/L2 files alone and compared the re-rendered text with the text it sent (\S\ref{sec:folding}); the check also verifies that L1 row ranges stay within their turn and that every tool output is covered by exactly one manifest row. Across the 809 folds of all LOCA-bench runs in this section, including the ablations and the sweep, none issued a model request and the check found one mismatch. It occurred in a zero-outline fold of a first run of variant D that was discarded for infrastructure errors and replaced by the run in Table~\ref{tab:ablation}; the other zero-outline folds passed.

The agent used the handles heavily after folds. In the main 1M runs it made 315 \ctx{} calls, 55 \scan{} calls, and 86 content-handle expansions; 21 of the \ctx{} calls (6.7\%) exited with a nonzero status.

\subsection{Case Study}
\label{sec:exp-case}

In a CanvasArrangeExam task of the fourth chain, the agent must build a final-exam schedule from course announcements and email. Early in the turn it fetched 160 courses (183\,KB, call \texttt{t8.13}), 34 emails (\texttt{t8.8}), and announcements across calls \texttt{t8.6}--\texttt{t8.177}, and saved the course table with \ctx{} as \texttt{derived/courses}. At 351K tokens the trigger folded the context to 83K in 928\,ms; the raw announcements left the window and remained as rows in the outline. After the fold, the agent listed turn 8 with \scan{}, then ran \ctx{} over the range \texttt{t8.6-177} (793\,KB, 16{,}708 lines) to ``extract final-exam status and details from every Canvas announcement call; one row per course'', receiving 6.4K tokens. It read the exam emails again through \texttt{t8.8}, joined the result with \texttt{derived/courses}, and searched both sources for exclusion signals before answering. The task passed. The 793\,KB of announcements never re-entered the window; the agent reached them through the call range the outline printed.

\section{Related Work}
\label{sec:related}

\paragraph{Organizing and compacting agent context.}
Observation masking drops old tool outputs~\citep{masking}; interaction-aware compression deletes old reasoning blocks while keeping actions and observations~\citep{forgetreasoning}; CliffCompaction truncates tool calls and outputs by rule once a budget is spent~\citep{cliff}. Generated and learned methods rewrite the history instead: ACON optimizes compression guidelines and distills them into a smaller compressor~\citep{acon}, and self-compaction, Context-Folding, and AgentFold train the agent to summarize sub-trajectories or fold its own workspace~\citep{selfcompact,contextfolding,agentfold}. In all of these the reduced view is produced when the reduction happens, and neither selection nor generation dominates in general~\citep{compactiontheory}. \folio{} produces its views during execution, so a fold selects among them.

\paragraph{Hierarchical and external memory.}
MemGPT pages between a bounded working context and external storage~\citep{memgpt}. HiAgent uses subgoals as memory chunks and replaces a finished subgoal's trajectory with a summarized observation~\citep{hiagent}. AgeMem trains an agent to call store, retrieve, update, summarize, and discard operations over long- and short-term memory~\citep{agemem}, and AgentRM manages context through a three-tier lifecycle inside an OS-style runtime~\citep{agentrm}.Agent Workflow Memory instead reuses routines across tasks~\citep{awm}. \folio{}'s levels differ in that each has a fixed author and role: the model writes only a short turn note, and the runtime writes the event manifest and the source records.

\paragraph{Recoverable context.}
Several systems keep the source reachable after reduction. LCM stores every message immutably and compacts the active context into a DAG of model-written summaries that tools can grep and expand~\citep{lcm}. SAM pairs raw trajectory pages with compact cues and has a trained memory model reread selected pages~\citep{sam}. SoL-Pi replaces large outputs with a handle and an excerpt and reduces logs to verified receipts~\citep{solpi}. ReCAP restores raw message blocks selected from a persistent attention graph~\citep{recap}, and CMV versions sessions as a DAG with snapshots and structural trimming~\citep{cmv}. RLM, CLM, and VISTA let the model operate on long inputs or on its own context through code, files, or a dashboard~\citep{rlm,clm,vista}. \folio{} shares the goal of recovery; its handles are the call coordinates of the fold itself, and recovery offers bounded reads, file-side computation, and a dossier over the same coordinates.

\paragraph{Cost under prompt caching.}
Measurements of prompt caching for agents show that cache-friendly context layouts cut API cost substantially~\citep{dontbreakcache}, and that fewer tokens need not mean lower cost once rewrites and re-acquisition are billed~\citep{tokencost,reacquisition}. Context Is Inventory derives restart intervals from an EOQ and ski-rental model~\citep{inventory}; CMV estimates how many turns a trim needs to repay its cache rebuild~\citep{cmv}; SoL-Pi's Online Context Compact compacts at plan-step boundaries when projected savings exceed the rewrite cost~\citep{solpi}. These models price when to reduce, but take the size after reduction as given by the reduction method.

\paragraph{Positioning.}
\folio{} is not a sum of a memory, a compactor, tools, and a trigger; the parts depend on each other through one handle space and one renderer. The levels exist before a fold, so the fold is a deterministic selection. Demotion is monotone, so the folded prefix stays cached. The fold prints the handles that L2 and L3 resolve, so recovery reads exactly the folded object and its cost lands in requests the trigger already accrues. The trigger takes the landing point from the renderer that builds the context, so the retained budget enters the cost decision. Table~\ref{tab:related} summarizes the comparison.

\begin{table*}[t]
\centering\small
\begin{tabularx}{\textwidth}{@{}lXccXc@{}}
\toprule
System & Reduced view & Model call to reduce & Source reachable & Access to folded content & Post-reduction size in trigger \\
\midrule
HiAgent & subgoal summaries & yes & yes & retrieve a subgoal trajectory & -- \\
LCM & DAG of summaries & yes & yes & describe, grep, expand & -- \\
CliffCompaction & rule-truncated history & no & re-run tools & tool signatures & fixed budget \\
SoL-Pi & handle + excerpt; log receipts; base summary & yes (compaction) & large outputs & paged handle read & no \\
Inventory & restart to working floor & -- & no & -- & fixed at floor \\
\folio{} & L1/L2 views formed during execution & no & yes & \scan{}, \ctx{}, \dossier{} on call handles & yes ($S=B+L+D$) \\
\bottomrule
\end{tabularx}
\caption{Comparison with the closest systems.}
\label{tab:related}
\end{table*}

\section{Conclusion}
\label{sec:conclusion}

\folio{} records a long agent history as three linked levels while the agent runs, folds it programmatically by selecting coarser views of existing rows, and lets the agent recover folded evidence through the handles the outline prints. Already sent lines are never redrawn richer, so the folded prefix stays cached, and the fold trigger prices an explicit landing point that includes the retained budget. On LOCA-bench it solves more tasks than Claude Code, Codex, SoL-Pi, and the same runtime with model-written compaction, in about half their time, and at 2.4 times lower cost than that runtime.

\section*{Limitations}

\paragraph{Evaluation scope.}
We evaluate one benchmark, LOCA-bench, with deterministic graders, and one model with one price structure. Each chain runs once per system, so the per-system comparison rests on 75 paired tasks; the closed systems expose no provider usage, so their cost is not compared. Variants C and D of the ablation and the landing sweep each cover one chain, and successes vary by 3--4 of 15 between chains of the same configuration. Tasks with open-ended outputs, or whose evidence can be cheaply re-queried from the environment, may benefit less from recovering captured history. Price ratios differ across providers, and the trigger's behavior depends on them.

\paragraph{Recoverability boundary.}
The contract of \S\ref{sec:problem} covers records that were captured, are still stored, and have valid handles. Content a tool never returned is outside it, and when a complete output file is missing or above 64\,MB, \scan{} returns the cut copy and says so. Recoverability does not imply that the agent finds or uses the right evidence.

\paragraph{Model-written state.}
L1 notes and dossier entries are written by the model and can be wrong or incomplete. An oversized dossier is consolidated by an auxiliary model request, which can drop or merge entries incorrectly.

\paragraph{Execution of \ctx{}.}
\ctx{} runs commands the model writes. Its isolation and permission model depend on the surrounding runtime, which this paper does not evaluate.

\paragraph{Fixed landing budget.}
We use one landing budget $L$ and report its sensitivity on one chain; we do not choose $L$ per task. Budgets above 75K were limited by the raw-area cap.

\bibliography{references}
\end{document}